\documentclass[final,5p,times,twocolumn]{elsarticle}

\pdfoutput=1

\usepackage{amssymb}
\usepackage{mathrsfs}

\usepackage{url}

\usepackage{algorithm}
\usepackage{algorithmic}

\usepackage{graphicx}
\usepackage{caption}
\usepackage[justification=centering]{caption}

\usepackage{subfigure}

\usepackage{amsmath}
\usepackage{cite}

\usepackage{array}
\usepackage{stfloats}

\usepackage{natbib}

\usepackage{threeparttable}
\usepackage{booktabs}
\usepackage{footnote}
\usepackage{ragged2e}
\usepackage{caption}
\usepackage[font=small,labelfont=bf,labelsep=none]{caption}
\usepackage{mathptmx}

\journal{Biomedical Signal Processing and Control}

\begin{document}

\begin{frontmatter}



\title{A Robust Deep Learning Approach for Automatic Classification of Seizures Against Non-seizures}


\author[label1]{Xinghua Yao}
\author[label1]{Xiaojin Li}
\author[label2]{Qiang Ye}
\author[label1]{Yan Huang}
\author[label1]{Qiang Cheng\corref{cor1}}
\ead{Qiang.Cheng@uky.edu}
\author[label3]{Guo-Qiang Zhang\corref{cor1}}
\ead{gqatcase@gmail.com}
\cortext[cor1]{Corresponding authors}

\address[label1]{Institute of Biomedical Informatics, University of Kentucky, Lexington, Kentucky, USA}
\address[label2]{Department of Mathematics, University of Kentucky, Lexington, Kentucky, USA}
\address[label3]{The University of Texas Health Science Center at Houston, Houston, Texas, USA}

\begin{abstract}
Identifying epileptic seizures through analysis of the electroencephalography (EEG) signal becomes a standard method for the diagnosis of epilepsy. Manual seizure identification on EEG by trained neurologists is time-consuming, labor-intensive and error-prone, and a reliable automatic seizure/non-seizure classification method is needed. One of the challenges in automatic seizure/non-seizure classification is that seizure morphologies exhibit considerable variabilities. In order to capture essential seizure patterns, this paper leverages an attention mechanism and a bidirectional long short-term memory (BiLSTM) to exploit both spatial and temporal discriminating features and overcome seizure variabilities. The attention mechanism is to capture spatial features according to the contributions of different brain regions to seizures. The BiLSTM is to extract discriminating temporal features in the forward and the backward directions. Cross-validation experiments and cross-patient experiments over the noisy data of CHB-MIT are performed to evaluate our proposed approach. The obtained average sensitivity of 87.00\%, specificity of 88.60\% and precision of 88.63\% in cross-validation experiments are higher than using the current state-of-the-art methods, and the standard deviations of our approach are lower. The evaluation results of cross-patient experiments indicate that, our approach has better performance compared with the current state-of-the-art methods and is more robust across patients.
\end{abstract}




\end{frontmatter}


\section{Introduction}
\label{section-introduction}
More than 50 million people in the world suffer from epilepsy \citep{Megiddo}. Epilepsy is a central nervous system disorder, in which brain activity becomes abnormal, causing seizures or periods of unusual behaviors, sensations, and sometimes loss of awareness.  An important technique to diagnose epilepsy is electroencephalography (EEG). An EEG signal records the electrical activities of the brain, and may reveal patterns of normal or abnormal brain electrical activities. In current clinical practices, EEG signals are collected from the brains by making use of either non-intrusive or implanted devices. The collected off-line EEG signals are then reviewed and analyzed by trained neurologists to identify characteristic patterns of the disease, such as pre-ictal spikes and seizures (A seizure is a sudden, uncontrolled electrical disturbance in the brain, which signifies epilepsy.), and to capture disease information, like seizure frequency, seizure type, etc. The obtained disease information is to provide supports for therapeutic decisions. This manual way of reviewing and analyzing is labor-intensive and error-prone, for it usually takes several hours for a well-trained expert to analyze one-day of recordings from one patient \citep{Gotman1979,Gotman1982,Thodoroff,Furbass,Zandi}. These limitations have motivated researchers to develop automated techniques to recognize seizure. In this paper, we focus on developing an automatic approach to classifying seizure signal segments and non-seizure segments from off-line EEG signals for assisting neurologists to make diagnosis.

One of critical challenges in the seizure/non-seizure classification is that seizure morphologies exhibit considerable inter-patient and intra-patient variabilities. Different machine learning methods and computational technologies have been applied to address this challenge. Seizure detection is often converted into a problem of seizure/non-seizure classification but more of a real-time flavor. Extensive studies have been coducted for constructing patient-specific detectors capable of detecting seizures \citep{Zandi,Shoeb,Amin,Fan,Hunyadi,Esbroeck,Truong2017,Vidyaratne,Qu,Zhou}. In early studies, hand-crafted features are usually used as characteristics of seizure manifestations in EEG. More recent studies focus on applying deep learning models to seizure detection \citep{Thodoroff,Vidyaratne,Golmohammadi,Acharya}. Most of these studies adopt interesting technologies to help extracting seizure features. For example, signal processing techniques are used to filer the data; certain modules need to be pre-trained; multiple channels are utilized to extract spatial features, and temporal features are extracted by the sliding windows. However, to the best of our knowledge, the data over channels are processed in the same way; i.e., the channels are not differentiated. About extracting temporal features, most studies only work in the forward direction. In fact, for seizure/non-seizure classification, the EEG signals can potentially provide some additional information in the backward direction \citep{Vidyaratne}.

Different brain regions are likely to have different contributions to the seizure. The characteristics of EEG data for epilepsy at different brain regions are different. The features of EEG signals at a time point are correlated with the past data and the future data. Besides, though EEG signals are in
general dynamic and non-linear, during a sufficiently small time period, the signal may be considered to be stationary. Based on the above three observations and inspired by an architecture in \citep{Hussein}, we design a new approach by using  bidirectional long short-term memory (BiLSTM) integrated with an attention mechanism. Firstly, we introduce an attention mechanism over EEG channels. Different weights are automatically assigned to signal channels at different brain regions according to how much they would affect the seizures. Secondly, the bidirectional long short-term memory  technique is adopted to extract temporal features of EEG signals in both the forward and the backward directions. Thirdly, output sequences of the BiLSTM module are split into patches according to time steps. Each patch only contains data at one time step. All the patches are separately processed to extract features. With these three new ideas, we develop a novel approach for seizure/non-seizure classification in EEG signals. Cross-validation and cross-patient experiments are performed using the proposed approach. In the cross-validation experiments, we obtain the average sensitivity, specificity and precision of 87.0\%, 88.6\% and 88.63\%, respectively, and the corresponding standard deviations of 0.0363, 0.0463 and 0.0388, respectively. For the cross-patient experiments, the average sensitivity, specificity and precision of 83.72\%, 84.06\% and 85.36\% are respectively achieved, and the standard deviations being 0.1349, 0.1379 and 0.1020, respectively. These results exceed the
current state-of-the-art performances on the noisy data of CHB-MIT in \citep{Acharya}, \citep{Hussein} and \citep{Thodoroff}. The extensive experimental results show that the performance of the proposed new approach is promising and has high stability, with smaller variations compared to existing methods.

In brief, the main novelties of our paper include the following:
\begin{itemize}
\item[(1)] An attention mechanism is utilized to capture spatial features of seizure for the first time. It distinguishes EEG signals from different brain regions and generates different attention weights for EEG data over different channels. The attention weights are explained by using EEG data segment examples.
\item[(2)] Bidirectional long short-term memory is combined with attention mechanism to extract temporal features. At each time step, the past spatially-weighted data and the future spatially-weighted data are analyzed.
\item[(3)] Experimental results on the noisy EEG data of CHB-MIT demonstrate that, the new approach can capture more robust seizure patterns than current state-of-the-art deep learning approaches, and overcome the inter-patient seizure variations better.
\end{itemize}

The rest of this paper is organized as follows. Section \ref{section-related-work} describes related research work on automatic seizure/non-seizure classification. Section \ref{section-methods} presents our designed approach of BiLSTM with attention. In Section \ref{section-evaluation}, evaluation of the proposed approach is performed in cross-validation and cross-patient experiments. Section \ref{section-model-analysis} explains the attention mechanism and validates main modules in the proposed approach. Section \ref{section-discussion} discusses the approach of BiLSTM with attention. Conclusions and future work are described in Section \ref{section-conclusion}.

\section{Related work}
\label{section-related-work}
There is extensive research for seizure/non-seizure classification, which distinguishes seizure segments from non-seizure segments. Seizure detection, which is often of a real-time flavor, is often viewed as the seizure/non-seizure classification problem. The study of seizure detection can be divided into three categories. One category is using traditional machine learning methods \citep{Shoeb,Amin,Hunyadi,Esbroeck,Truong2017,Fergus,Nicalaou,Nasehi,Kharbouch,Zheng}. The second category is about signal processing methods and network techniques \citep{Zandi,Fan,Zhou,Acharya2012}. And the third category is using deep learning methods \citep{Thodoroff,Vidyaratne,Golmohammadi,Acharya,Hussein,Ansari}.

\subsection{Work based on machine learning methods}
With traditional machine learning methods, many previous works focus on developing patient-specific seizure detection methods.

Shoeb and Guttag proposed a patient-specific seizure detection method by using the support vector machine (SVM) \citep{Shoeb}. The method leverages filters to extract spectral features over each channel, and then concatenate the feature vectors according to a fixed time length. Then, train the SVM model with the obtained feature vectors as the input. The method achieved a sensitivity of 96\%, a median detection delay of 3 seconds and a median false detection rate of 2 per 24 hour. The sensitivity result is often used as a benchmark for patient-specific seizure detection on the data set CHB-MIT. The authors observed that the identity of channels could help differentiate between the seizure and the non-seizure activity.

Amin and Kamboh \citep{Amin} designed an algorithm RUSBoost to process imbalanced seizure/non-seizure data, and used RUSBoost and the decision tree classifier to conduct patient-specific experiments with the CHB-MIT data set. The method was fast in training and achieved good performance with seizure detection accuracy of 97\% and false detection rate of 0.08 per hour.

Hunyadi et al. \citep{Hunyadi} presented seizure detection algorithm, which uses a nuclear norm regularization to convey spatial distribution information of ictal patterns. The algorithm extracted features from each channel, and then stacked them to analyze as one entity.

Truong et al. \citep{Truong2017} proposed a automatic seizure detection method over intracranial electroencephalography (iEEG) data.
First, supervised classifiers were used to select those channels that contribute the most to seizures. Features in the frequency and time domains were extracted, including spectral power and correlations between channel pairs. Then, Random Forest classifier was utilized for classification. This method has the state-of-the-art computational efficiency while maintaining the accuracy. In this method, selecting channels that contribute the most to seizures is to reduce the number of channels, thereby improving the computational efficiency.

The work in \citep{Shoeb,Amin,Hunyadi,Truong2017} used data over multiple channels to extract spatial features. However, they did not apply different processing ways to the data with different channels.

Esbroeck et al. \citep{Esbroeck} proposed a multi-task learning framework to detect patient-specific seizure onset in the presence of intra-patient variability in seizure morphology. They considered distinguishing the windows of each seizure from non-seizure data as a separate task and treating the individual-seizure discrimination as another task. Compared to the standard SVM, testing results of the CHB-MIT data set indicated that their approach performed better in most cases.

Kiranyaz et al. \citep{Kiranyaz} presented a systematic approach for patient-specific classification of long-term EEG. In the approach, EEG data were processed through band-pass filtering, feature extraction, epileptic seizures aggregation and morphologic filtering. Results of the data processing were input into collective network of binary classifiers to classify signal from each channel. Then, initial classification results over each channel were further learned and weighted by a dedicated classifier which makes final classification decision of each EEG frame. Over the CHB-MIT data set, \citep{Kiranyaz} achieved an average sensitivity of 89.01\% and an average specificity of 94.71\%. High number of classifiers increased computational complexity of the approach.

In the patient-specific case, the data have no variations caused by different subjects. The performances of the patient-specific seizure/non-seizure classifiers are better than 90\%. However, the patient-specific classifiers have a limitation of poor generalizability.

In \citep{Fergus}, Fergus et al. presented a method for seizure detection across subjects based on traditional machine learning techniques, and obtained
88\% in Sensitivity and 88\% in Specificity over the CHB-MIT data set by selecting features in multiple brain regions. The method mainly consists of four steps, which are data filtering, feature extraction, feature selection and training classifiers. In cross-validation experiments, EEG signals in CHB-MIT were segmented according to a segment length 60 seconds, one seizure segment was truncated for each seizure, non-seizure segments were extracted from non-seizure EEG records as many as seizure segments. The produced experiment data consist of 171 seizure segments and 171 non-seizure segments. On the average, each seizure segment contains 40s seizure data. Additionally, after segmenting EEG signals \citep{Fergus} used a bandpass filter and second order butterworth filters to extract the EEG data in the bandwidth 0.5Hz-30Hz.

\subsection{Work based on signal processing and network techniques}
Based on signal processing techniques, Zandi et al. proposed a wavelet-based algorithm for real-time detection of epileptic seizures using scalp EEG \citep{Zandi}. In this algorithm, the EEG from each channel was decomposed by wavelet packet transform, and a patient-specific measure was developed by using wavelet coefficients to separate the seizure and non-seizure states. Utilizing the measure, a combined seizure index was derived for each epoch of every EEG channel. Appropriate channel alarms were generated by inspecting the combined seizure index.

Acharya et al. \citep{Acharya2012} presented a method for the automatic detection of normal, pre-ictal, and ictal conditions from EEG signals. Four entropy features, including approximate entropy, sample entropy, and two phase entropies, were extracted. The extracted features were input into the classifier to do classification. Over the EEG data set provided by University of Bonn, seven classifiers were fed with extracted entropies to show the effectiveness of the features.

Zhou et al. \citep{Zhou} proposed a seizure detection algorithm using lacunarity and Bayesian linear discriminant analysis (BLDA). In the algorithm, wavelet decomposition on EEGs was conducted with five scales, and the wavelet coefficients at scales 3, 4, and 5 were selected. Features including lacunarity and fluctuation index were extracted from the selected scales, and then they were fed to the BLDA for training and classification. Patient-specific experiments were performed on intracranial EEG data from the Epilepsy Center of the University Hospital of Freiburg. The obtained average sensitivity was 96.25\%, with an average false detection rate of 0.13 per hour and a mean delay time of 13.8s. The obtained precision results for eleven patients were less than 50\%.

By leveraging network technologies, Fan and Chou \citep{Fan} utilized a complex network model to represent EEG signals, and integrated it with spectral graph theory to extract spatial-temporal synchronization patterns for detecting seizure onsets in real-time. The method was tested on 23 patients from the CHB-MIT data set. The resulting patient-specific sensitivity surpassed the benchmark methods.

\subsection{Work based on deep learning methods}
Recently, deep learning techniques have been developed rapidly and applied to solve the seizure/non-seizure classification problem.

Vidyaratne et al. \citep{Vidyaratne} proposed a deep recurrent architecture by combining Cellular Neural Network and Bidirectional
Recurrent Neural Network. The bidirectional recurrent neural network was deployed into each cell in the cellular neural network, and it was utilized to extract temporal features in the forward and the backward directions. Each cell interacts with its neighbor cells to extract local spatial-temporal features. The computed results in the cellular neural network were output into a multi-layered perceptron. In the perceptron, samples
 were classified based on a trained threshold. In order to satisfy the input requirements of cellular neural network, the authors proposed a mapping which organizes EEG signals into a 2D grid arrangement. Patient-specific experiments were conducted over the EEG data of five patients from the CHB-MIT data set. The obtained sensitivities were all 100\% for the five patients. In their experiments, the raw EEG data were preprocessed using a bandpass filter between 3Hz and 30Hz in order to extract seizure activity data.

Golmohammadi et al. \citep{Golmohammadi} explored seizure-detection performances of two neural networks over the data source of TUH EEG Corpus. Their experiment results showed that the convolutional long short-term memory (LSTM) network is better than the convolutional GRU network. And also the impacts of initialization methods and regularization methods over the performance were experimented. The two models in \citep{Golmohammadi} did not utilize attention mechanism.

Hussein et al. \citep{Hussein} designed a deep neural network for seizure/non-seizure classification by using LSTM as a main module. The approach extracts temporal features by using LSTM. Evaluation was performed on the EEG data set provided by University of Bonn. Testing results mostly reached 100\%. In \citep{Acharya}, Acharya et al. presented a 13-layers deep neural network for seizure/non-seizure classification by using convolutional neural network (CNN). Over the Bonn EEG data set, the obtained average sensitivity and specificity were 95\% and 90\%, respectively. For the experiments in \citep{Hussein} and \citep{Acharya}, the two approaches extracted seizure features from the data on one channel to conduct classification. Each record in the Bonn EEG data set is the data from only one channel.

In \citep{Thodoroff}, Thodoroff et al. designed a recurrent convolutional neural network to capture spectral, spatial and temporal patterns of
seizures. The EEG signals were firstly transformed into images by using Polar Projection, cubic interpolation, and Fast Fourier transform. The image-based representation of EEG signals was to exploit the spatial locality in seizures. Created images were fed to the convolution
neural network. The output vectors of the convolution neural network were organized to be sequences in chronological order. The sequences were then input into the bidirectional recurrent neural network to produce classified seizure/non-seizure results. Both patient-specific experiments and cross-patient experiments were performed. The patient-specific experiment results were similar to the results in \citep{Shoeb}. And the cross-patient testing sensitivity was 85\% on average. In the two kinds of experiments, the convolution neural network was pre-trained alone. And the transfer learning technology was utilized to overcome the problem of small amount of data in the patient-specific experiments. The proposed recurrent convolutional neural network in \citep{Thodoroff} is complicated.

Ansari et al. \citep{Ansari} aimed to automatically optimize feature selection for seizure detection. They utilized deep CNN to extract optimal features, and then fed the features to random forest to do classification. In evaluation experiments, EEG recordings of 26 and 22 neonates were taken as training data and testing data, respectively. A false alarm rate of 0.9 per hour and a sensitivity of 77\% were achieved. The proposed method needed no predefined features, and surpassed three classic feature-based approaches.

\section{Methods}
\label{section-methods}

\subsection{Model design}
EEG signal data is an important modality for the diagnosis of epilepsy. It is generally collected through placing electrodes on the scalp. Each electrode records brain activities in its located brain region. As different brain regions play different roles in the seizure procedure, the
data collected at different brain regions record different characteristics of seizures. With the observations in \citep{Shoeb}, differences between seizure data and non-seizure data are related to channels. To exploit the differences of signals from different brain regions, we will use an attention mechanism to assign different weights to data from different channels.

Brain activities are continuous, and EEG signals could be regarded as continuous records of brain activities when ignoring the sampling effects. The brain activity at a time point is correlated with past signal data, and could also be analyzed from future signal data. This two-directions analysis helps extract more seizure features. To leverage correlations from both directions, we perform BiLSTM for analyzing EEG sequence data.

EEG signal is dynamic and non-linear. Due to the dynamic nature, certain statistical characteristics of EEG signals change over time. However, the EEG signal segments have similar statistical temporal and spectral features for a sufficiently small time duration \citep{Hussein,Hassanpour}. After bidirectionally processing, the sequence is split into time-step patches. Each patch only contains data in a time step. The patches are further extracted
features through full connection operations separately and concurrently.

Based on the above three ideas and inspired by \citep{Hussein}, we develop a new approach of BiLSTM with attention (shortly, attention BiLSTM) in order to classify seizure segments and non-seizure segments. Raw EEG signals are split into data segments according to a fixed time span. The split data segments are automatically weighted through an attention mechanism, i.e., for each segment, signal data from different channels are
multiplied with different weights. The weights are achieved through a fully connected module and a non-linear function in training procedure. After adding weights, the data segments are fed to bidirectional LSTM module. The BiLSTM module extracts features in both forward and backward directions.
For output sequences of BiLSTM, data at each time step are separately input into a full connection module. Then, the extracted features are averaged over all the time steps in order to achieve global features of a segment. Finally, the labels of data segments are calculated by a fully connected module with the Softmax function.

\subsection{Model architecture and algorithm}
\begin{figure}[htb]
\centering
\includegraphics[scale=0.35]{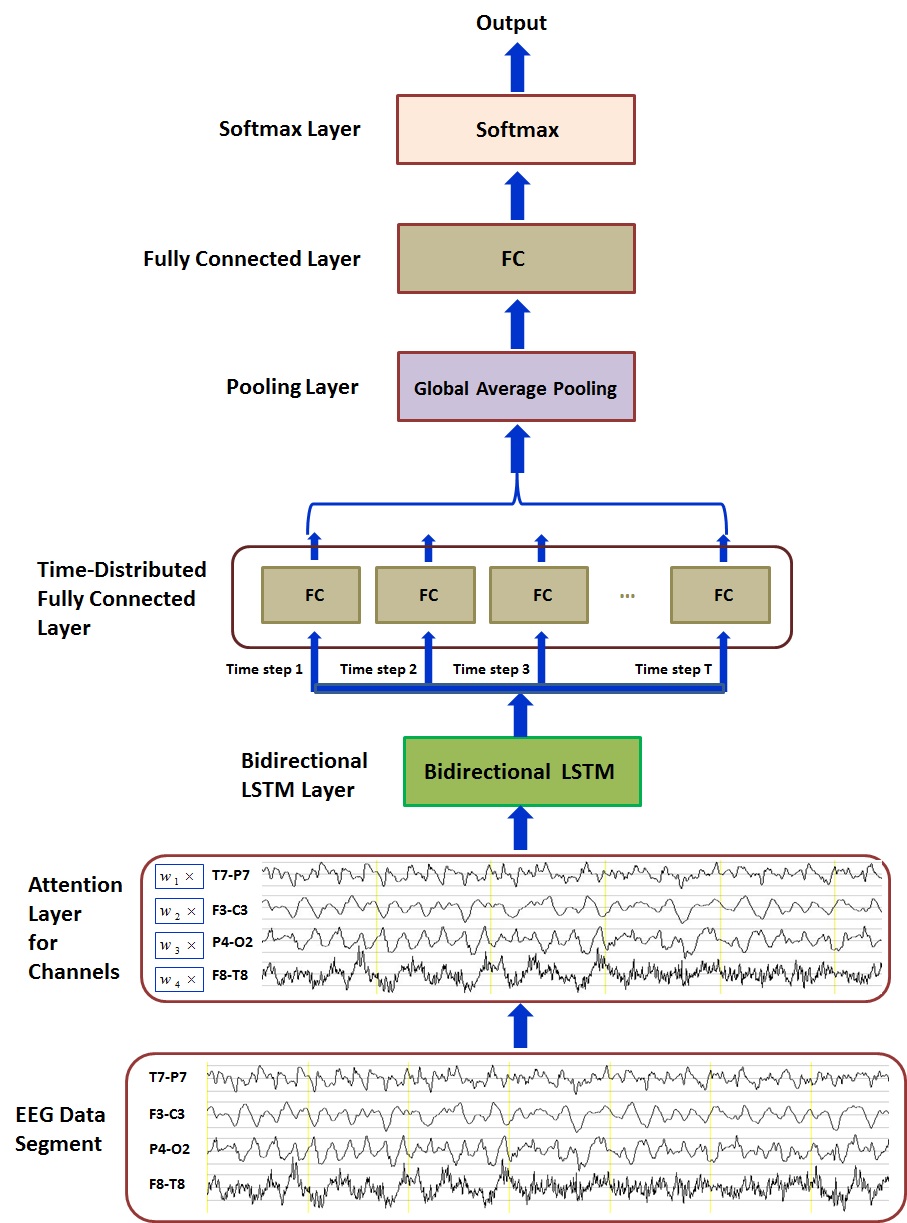}
\caption{Architecture of the proposed approach. T7-P7, F3-C3, P4-O2 and F8-T8 represent channels. $W_{1}$, $W_{2}$, $W_{3}$ and $W_{4}$ are weights on the four channels, respectively.}
\label{fig-model-architecture}
\end{figure}

Our model architecture consists of five modules, including attention layer, BiLSTM module, time-distributed fully-connected layer, pooling layer and fully-connected layer with Softmax. The designed architecture is presented in Fig. \ref{fig-model-architecture}.

\subsubsection{Attention layer}
\begin{figure}[htb]
\centering
\includegraphics[scale=0.37]{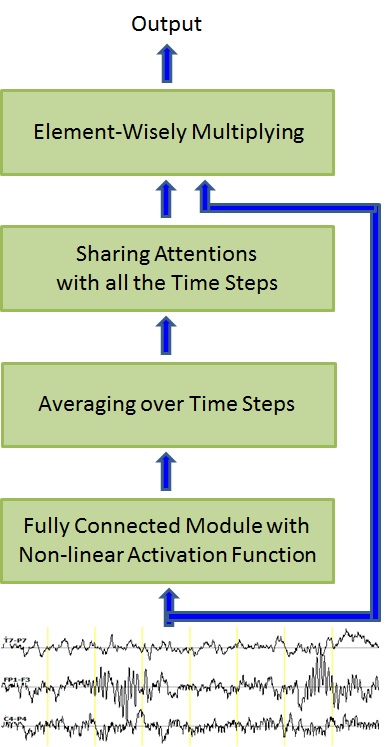}
\caption{Work flow of attention layer.}
\label{fig-attention-layer}
\end{figure}

The attention layer, shown in Fig. \ref{fig-attention-layer}, is to generate attention weights for each channel and then executes an element-wise multiplication. The original data are input into a fully connected module with a nonlinear activation function. The outputs of the fully connected
module are averaged over all the time steps. Then, the obtained average values are copied to be shared at all time steps. In this way, an attention weight matrix is achieved. Finally, the attention matrix is element-wisely multiplied with the original inputs. The attention layer is computed using the following equations:
\begin{align}
\label{eq-atten-layer-reshape-1} Y_{1}  & =  f_{re_{1}}(X_{0})  \\
\label{eq-atten-layer-fully-connected} Y_{2} & =  \sigma(Y_{1} * W_{al} + B_{al}) \\
\label{eq-atten-layer-reshape-2} Y_{3} & = f_{re_{2}}(Y_{2})  \\
\label{eq-atten-layer-average} Y_{4} & = f_{av}(Y_{3}) \\
\label{eq-atten-layer-copy} Y_{5} & = f_{cy}(Y_{4})  \\
\label{eq-atten-layer-element-wise-multi} Y_{al} & = X_{0}\odot Y_{5}
\end{align}
Here, $X_{0}$ denotes an input tensor of size $(n_{sm},n_{sp},n_{ch})$. Symbols $n_{sm}$, $n_{sp}$, $n_{ch}$ represent the number of samples, the number of time steps, and the number of signal channels, respectively. $Y_{1}$ is a matrix of size $(n_{ss},n_{ch})$, $n_{ss}=n_{sm}*n_{sp}$, $W_{al}$ a weight matrix of size $(n_{ch},n_{ch})$, a bias matrix $B_{al}$ of size $(n_{ss},n_{ch})$, and $Y_{2}$ with size $(n_{ss},n_{ch})$. A symbol $\sigma(\cdot)$ represents a non-linear function, like $softmax(\cdot)$ and $sigmoid(\cdot)$. $Y_{3}$ is a matrix of size $(n_{sm},n_{sp},n_{ch})$, $Y_{4}$ of size $(n_{sm},n_{ch})$, $Y_{5}$ of size $(n_{sm},n_{sp},n_{ch})$, and $Y_{al}$ an output matrix of attention layer with shape $(n_{sm},n_{sp},n_{ch})$. Functions $f_{re_{1}}(\cdot)$ and $f_{re_{2}}(\cdot)$ are to reshape a matrix, $f_{av}(\cdot)$ is a function of computing averages along with the second axis of matrix, and $f_{cy}(\cdot)$ is an copying operation to share the averages over all the time steps. The symbol $\odot$ means an element-wise multiplication between matrices.

\subsubsection{BiLSTM module}
\begin{figure}[htb]
    \centering
    \includegraphics[scale=0.32]{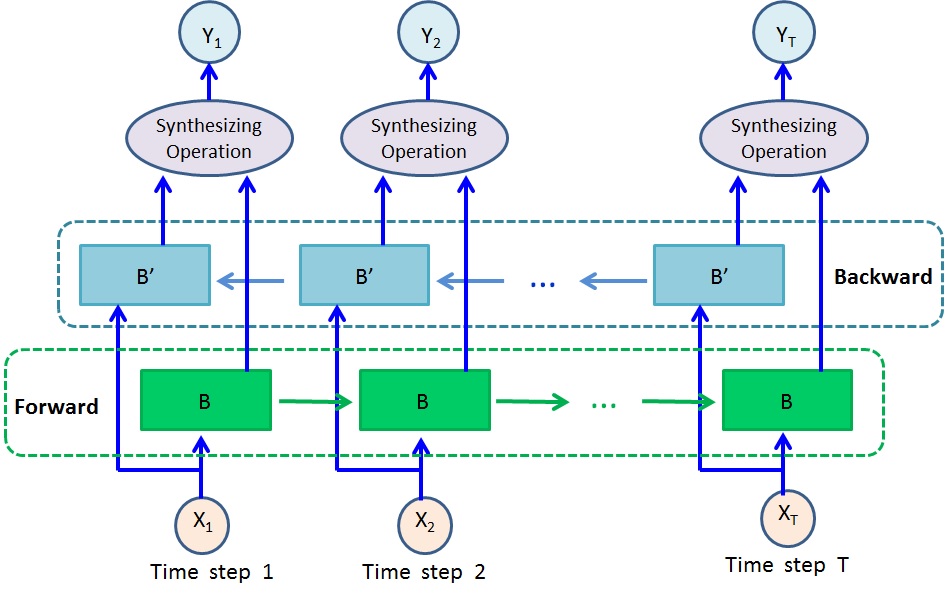}
    \caption{Work flow of BiLSTM module.}
    \label{fig-BiLSTM-module}
\end{figure}

The BiLSTM module processes the input sequence separately according to the forward order and the backward order, and synthesize the forward outputs and the backward outputs \citep{Schuster,Graves}. Its main procedure is presented in Fig. \ref{fig-BiLSTM-module}. In either forward order or backward order, the sequence is computed in the same way as LSTM, in which the computation can be described by using Eqs. (\ref{eq-block-input})$-$(\ref{eq-block-output}) according to \citep{Gers} and \citep{Greff}. The synthesizing operations can be concatenation or summation.
\begin{align}
    \label{eq-block-input} \text{Block input} & & \widetilde{C}_{t} & = \varphi(X^{in}_{t} * W_{ce} + Y^{bo}_{t-1} * R_{ce} + B_{ce})  \\
    \label{eq-input-gate} \text{Input gate} & & G^{ig}_{t} & = \sigma(X^{in}_{t} * W_{ig} + Y^{bo}_{t-1} * R_{ig} + B_{ig}) \\
    \label{eq-forget-gate} \text{Forget gate} & & G^{fg}_{t} & = \sigma(X^{in}_{t} * W_{fg} + Y^{bo}_{t-1} * R_{fg} + B_{fg}) \\
    \label{eq-output-gate} \text{Output gate} & & G^{og}_{t} & = \sigma(X^{in}_{t} * W_{og} + Y^{bo}_{t-1} * R_{og} + B_{og})\\
    \label{eq-cell} \text{Cell} & & C_{t} & = C_{t-1}\odot G^{fg}_{t} + \widetilde{C}_{t}\odot G^{ig}_{t}  \\
    \label{eq-block-output} \text{Block output} & & Y^{bo}_{t} & = \psi(C_{t})\odot G^{og}_{t}
\end{align}
Here, $X^{in}_{t}$ is an input matrix of size $(n_{sm},n_{ch})$ at the time step $t$, and $Y^{bo}_{t}$ an output matrix of size $(n_{sm},n_{fe_{1}})$ at the time step $t$, where $n_{fe_{1}}$ is a dimensionality of extracted feature space. Matrices $G^{ig}_{t}$, $G^{fg}_{t}$, $G^{og}_{t}$, $\widetilde{C}_{t}$, and $C_{t}$ represent input gate state, forget gate state, output gate state, a block input, and cell state at the time step $t$, respectively. Input weights matrices $W_{ce}$, $W_{ig}$, $W_{fg}$ and $W_{og}$ are with shape $(n_{ch},n_{fe_{1}})$. Recurrent weights matrices $R_{ce}$, $R_{ig}$, $R_{fg}$, and $R_{og}$ are of size $(n_{fe_{1}},n_{fe_{1}})$. Bias matrices $B_{ce}$, $B_{ig}$, $B_{fg}$, and $B_{og}$ are of size $(n_{sm},n_{fe_{1}})$. $\varphi(\cdot)$, $\sigma(\cdot)$, and $\psi(\cdot)$ are non-linear activation functions. The symbol $\odot$ means element-wise multiplication.

For the output $Y_{al}$ in the attention layer, it is split into $n_{sp}$ components according to time steps, i.e., $X_{1}, X_{2}, \cdots, X_{n_{sp}}$, with each one being a matrix of size $(n_{sm},n_{ch})$. These components form a sequence of $X_{1}X_{2}\cdots X_{n_{sp}}$ in a chronological orders. For the sequence $X_{1}X_{2}\cdots X_{n_{sp}}$, the variable $X_{in_{t}}$ in Eq. (\ref{eq-block-input}) has different values in the forward and the backward order. Its value at the time step $t$ in the forward order is $X_{t}$, and the value in the backward order is $X_{n_{sp}-t+1}$. Based on Eqs. (\ref{eq-block-input})$-$(\ref{eq-block-output}), a forward output sequence $\mathscr{Y}_{fd}$ is obtained in the forward order, and a backward output sequence $\mathscr{Y}_{bd}$ for the backward order. We use $\mathscr{Y}_{fd}(t)$ to denote the $t$-th item in the sequence $\mathscr{Y}_{fd}$, i.e., the forward output at the time step $t$, and $\mathscr{Y}_{bd}(t)$ for the backward output at the time step $t$. The two output sequences $\mathscr{Y}_{fd}$ and $\mathscr{Y}_{bd}$ are then synthesized as follows:
\begin{equation}
\label{eq-synthesize-biLSTM} \mathscr{Y}_{blm}(t)=\Phi(\mathscr{Y}_{fd}(t),\mathscr{Y}_{bd}(n_{sp}-t+1))
\end{equation}
Here, $t=1,\cdots,n_{sp}.$ $\Phi(\cdot)$ means an operation, which has two options, i.e., concatenation and summation. $\mathscr{Y}_{blm}$ represents the synthesized sequence of the forward output sequence and the backward output sequence, and $\mathscr{Y}_{blm}(t)$ of size $(n_{sm},n_{fe_{2}})$ means the $t$-th item in the sequence $\mathscr{Y}_{blm}$, i.e., the output of BiLSTM module at the time step $t$. $n_{fe_{2}}$ is a dimensionality of output space of BiLSTM module.

\subsubsection{Time-distributed fully-connected layer}
The time-distributed fully-connected layer is to further extract features at each time step. It executes fully-connected operations separately and simultaneously for inputs at each time step. And the fully-connected operations use linear functions as activation
functions. Time-distributed layer could help improve executing efficiency when processing signal data with high sampling
frequency. At each time step, the computation procedure is described as follows:
\begin{equation}
\label{eq-time-distributed-FC} \mathscr{Y}_{dl}(t)=\mathscr{Y}_{blm}(t)*W_{dl}+B_{dl}.
\end{equation}
Here, $t=1,2,\cdots,n_{sp}$. Matrix $\mathscr{Y}_{dl}(t)$ of size $(n_{sm},n_{fe_{3}})$, is the output at the time step $t$ in time-distributed fully-connected layer, where $n_{fe_{3}}$ is a dimensionality of extracted feature space in the time-distributed layer. $W_{dl}$ denotes a weight matrix of size $(n_{fe_{2}},n_{fe_{3}})$, $B_{dl}$ a bias matrix of size $(n_{sm},n_{fe_{3}})$. All the time-step components $\{\mathscr{Y}_{dl}(t), \ t=1,\cdots,$ $n_{sp}\}$ compose a matrix $Y_{dl}$ of size $(n_{sm},n_{sp},n_{fe_{3}})$ as the output of the time-distributed fully-connected layer.

\subsubsection{Pooling layer}
The pooling layer in our architecture executes the average pooling operation in order to extract global features of each sample. The operation computes a mean value of the time-step data for each sample in the output matrix $Y_{dl}$ of time-distributed fully-connected layer, and outputs a matrix $Y_{ap}$ of size $(n_{sm},n_{fe_{3}})$.

\subsubsection{Fully connected layer and Softmax layer}
Fully connected layer executes a fully connected operation to extract further features and to reduce the last dimension of input
matrix into number of classes. It uses a linear function as its activation function. Based on outputs of the fully-connected layer, Softmax layer computes probabilities that each sample belongs to a classification. In the following, we will use Eq. (\ref{eq-fully-connected}) and Eq. (\ref{eq-softmax}) to present the computations in the fully-connected layer and in the Softmax Layer.
\begin{align}
\label{eq-fully-connected}  Y_{fcl} & = Y_{ap}*W_{fcl}+B_{fcl} \\
\label{eq-softmax}  Y_{sl} & = softmax(Y_{fcl})
\end{align}
Here, $W_{fcl}$ and $B_{fcl}$ denotes weights matrix of size $(n_{fe_{3}},n_{c})$ and bias matrix of size $(n_{sm},n_{c})$, respectively. $n_{c}$ is the number of classes. $Y_{fcl}$ is an output matrix of size $(n_{sm},n_{c})$ in the fully-connected layer. Function $softmax(\cdot)$ calculates probabilities about each sample belonging to each class. $Y_{sl}$ is an output of the Softmax layer.

The pseudo-codes of the proposed seizure/non-seizure classification approach of BiLSTM with attention are shown in Algorithm 1.
\begin{algorithm}[htb]
\small
\caption{Seizure/Non-seizure Classification over EEG Data using the Attention BiLSTM Approach}
\label{algorithm}
\begin{algorithmic}[1]
\REQUIRE $X_{0}$, the matrix of EEG data segments
\ENSURE $Y_{pred}$, the matrix of classification results
\STATE Initialize matrices $W_{al}$, $B_{al}$, $W_{ce}$, $W_{ig}$, $W_{fg}$, $W_{og}$, $R_{ce}$, $R_{ig}$, $R_{fg}$, $R_{og}$, $B_{ce}$, $B_{ig}$, $B_{fg}$, $B_{og}$, $W_{dl}$, $B_{dl}$, $W_{fcl}$, $B_{fcl}$
\STATE Compute the output matrix $Y_{al}$ using the input $X_{0}$ and Eqs. (\ref{eq-atten-layer-reshape-1})$-$(\ref{eq-atten-layer-element-wise-multi})
\STATE Split $Y_{al}$ into $n_{sp}$ components $\{X_{1}, X_{2}, \cdots, X_{n_{sp}}\}$ according to time steps, and compose a sequence $X_{1}X_{2}\cdots X_{n_{sp}}$ in chronological order
\STATE Compute a forward output sequence $\mathscr{Y}_{fd}$ for the sequence $X_{1}X_{2}\cdots X_{n_{sp}}$ based on Eqs. (\ref{eq-block-input})$-$(\ref{eq-block-output})
\STATE Compute a backward output sequence $\mathscr{Y}_{bd}$ for the inverse sequence $X_{n_{sp}}\cdots X_{2}X_{1}$ based on Eqs. (\ref{eq-block-input})$-$(\ref{eq-block-output})
\STATE Synthesize sequences $\mathscr{Y}_{fd}$ and $\mathscr{Y}_{bd}$ by using Eq. (\ref{eq-synthesize-biLSTM}), and achieve a sequence $\mathscr{Y}_{blm}$
\STATE Compute a sequence $\mathscr{Y}_{dl}$ by using Eq. (\ref{eq-time-distributed-FC}), and then compose a matrix $Y_{dl}$ according to time steps
\STATE Compute matrix $Y_{ap}$ by averaging values over time steps for each sample in $Y_{dl}$
\STATE Compute matrix $Y_{sl}$ according to Eqs. (\ref{eq-fully-connected}) and (\ref{eq-softmax})
\STATE Compute the column position of the maximal element in each row of $Y_{sl}$, and achieve classification results $Y_{pred}$
\STATE Return $Y_{pred}$

\end{algorithmic}
\end{algorithm}

\section{Evaluation}
\label{section-evaluation}
In this section, we evaluate the approach of BiLSTM with attention by performing cross-validation experiments and cross-patient experiments over the noisy scalp EEG data set of CHB-MIT. Our evaluation mainly adopts three standard metrics, including the sensitivity, the specificity and the precision. The cross-validation experiment is that, data from all the patients are randomly split into three mutually disjoint sets, i.e., training set, validation set and testing set. The training set and validation set are used to train a model, and the testing set is to assess the ability of the trained model. To reduce variability, ten rounds of cross-validation are performed for each seizure/non-seizure classification approach in our experiments. Then, average values and standard deviations over results in the ten rounds are calculated. The cross-patient experiment means that, one patient is selected as testing subject, and all the other patients as training and validation subjects. Data from the training and validation subjects are to train a model, and data from the testing subject are to test the trained model. In our cross-patient experiments, 23 patients in CHB-MIT are separately selected as test subject to assess the performance of our proposed approach, and then the overall performance over the 23 patients is analyzed.

\subsection{Data}
\subsubsection{CHB-MIT data set}
The data set CHB-MIT contains 686 EEG recordings from 23 patients of different ages ranging from 1.5 years to 22 years. The recordings include
198 seizures. The used sampling frequency is 256 Hz. Each recording contains a set of EEG signals with different channels. Most recordings are one hour long, and some are for two or four hours. The EEG recordings are grouped into 24 cases and stored in EDF data files. Each EDF file corresponds to an EEG recording. In each case, the signal data were recorded from a single patient. Case Chb21 was obtained 1.5 years after Case Chb01 from the same patient. Each data file contains data over 23 or more channels. There exist data files in which the data over some channels were missing. And some data files, for example, Chb12\_27.edf, Chb12\_28.edf and Chb12\_29.edf, have different channel montages from other seizure files. In our experiments, we did not use the data in the above three EDF files.

\subsubsection{Data segmentation}
In order to extract effective seizure features, 17 common channels were selected, i.e., for each patient, the data of 17 common channels
 were used for seizure/non-seizure features extraction. The 17 common channels were P4-O2, FP2-F4, P7-O1, C4-P4, F7-T7, C3-P3, FP1-F7, F8-T8, FZ-CZ, CZ-PZ, F3-C3, T7-P7, P8-O2, FP1-F3, F4-C4, FP2-F8, and P3-O1, respectively. Each data record was split into data segments with the length of 23 seconds from the beginning to the end without overlapping. According to annotation files which mark the starting time and the ending time of each seizure, it could be determined whether a data segment contains a seizure or not. In our experiments, if a segment contained a seizure, it was considered as a seizure segment; otherwise, it was a non-seizure segment. In the seizure segments, the lengths of seizure data varied from 1s to 23s, with the average length being 16.9s. Among all seizure segments, the portion of the seizure signal less than 7s was 14.7\%, the part containing more than 10s accounted for 76.1\%, and the part containing more than 17s accounted for 59.8\%.

As a result of the splitting, 665 seizure segments were obtained. The 665 seizure data segments were taken as a part of our experiment data. For evaluation over a balanced data, 665 non-seizure segments in each experiment were randomly selected from all the non-seizure segments.

\subsection{Cross-validation seizure/non-seizure classification}
The deep learning approach in \citep{Hussein} uses LSTM as a main module (shortly, LSTM approach) to detect seizures. The LSTM approach is evaluated through cross-validation experiments over the EEG data set from University of Bonn \citep{Andrzejak}, showing the state-of-the-art performance. We will compare our approach with the LSTM approach. And also our approach will be compared with a convolutional neural network approach (for short, CNN approach) in \citep{Acharya}. Since the data in Bonn EEG data set is strictly processed, and does not contain any artifacts, and is small in size, we choose to use the noisy CHB-MIT data set for the cross-validation experiments.

The LSTM approach \citep{Hussein} and the CNN approach \citep{Acharya} do not provide all the source codes. Thus, we implemented the two
approaches according to their descriptions. The implemented LSTM approach and CNN approach were tested, and the testing results achieved the reported performances in \citep{Hussein} and \citep{Acharya}. Then based on the two implementations, we experimented with the CHB-MIT data set to compare them with our approach of attention BiLSTM.

In each cross-validation experiment, all the seizure segments were utilized as a part of experiment data, and non-seizure segments with the same quantity were randomly selected. The training set, validation set and testing set were obtained by randomly splitting the experiment data set according to the ratio 70:15:15. We tuned and determined parameters to achieve the best performance for the three approaches, including the LSTM approach, the CNN approach, and our attention BiLSTM approach. And for each approach, ten cross-validation experiments were carried out based on the correspondingly well-tuned parameters.

For cross-validation experiments using the LSTM approach, our parameters were set as follows: The number of hidden states was 120 in the LSTM layer, that in the time-distributed computing layer was 60, the optimizer was RMSprop, the learning rate was 0.0007, the batch size was 30, and the number of epochs was 30. For the CNN approach in \citep{Acharya}, it contains five convolutional layers, five max pooling layers, and three fully connected layers, and its parameters setting in our cross-validation experiments was as follows: The number of hidden states in the first two convolutional layers was 100, that in each of the second two convolutional layers was 200, that in the fifth convolutional layer was 260, that in the first fully connected layer was 100, that in the second fully connected layer was 50, the parameter alpha was 0.01 in the LeakyReLU activation function, the optimizer was Adam, the learning rate was 0.001, the batch size was 30, and the number of epochs was 50. For the proposed approach of BiLSTM with attention, our well-tuned parameters in the cross-validation experiments were as follows: The number of hidden states in the bidirectional LSTM layer was 140, that in the time-distributed layer was 70, the merging mode in the bidirectional LSTM was concatenation, the optimizer was RMSprop, the learning rate was 0.0013, the batch size was 30, and the number of epochs was 35. And the total number of trainable parameters is 197,078.

The cross-validation results using the LSTM approach, including Sensitivity, Specificity, F1 score, Precision, Accuracy, the average and the standard  deviation, are shown in Table \ref{tab-mixing-patients-results-LSTM}. And the results by using the CNN approach and our approach of attention BiLSTM are presented in Tables \ref{tab-mixing-patients-results-CNN} and \ref{tab-mixing-patients-results-BiLSTM}, respectively.
\begin{table}[!htb]
\centering
\fontsize{6}{8}\selectfont
\setlength{\abovecaptionskip}{0pt}
\setlength{\belowcaptionskip}{5pt}
\caption{Cross-validation results using the LSTM approach.}
\label{tab-mixing-patients-results-LSTM}
\resizebox{\columnwidth}{!}{
\begin{tabular}{l l l l l l}
\specialrule{0.3pt}{1pt}{2pt}
Item & Sens. & Spec. & F1 Sco. & Prec. & Accu. \\
\specialrule{0.2pt}{2pt}{1pt}
1 & 0.8500 & 0.8800 & 0.8629 & 0.8763 & 0.8650 \\
2 & 0.7700 & 0.8500 & 0.8021 & 0.8370 & 0.8100 \\
3 & 0.7900 & 0.8700 & 0.8229 & 0.8587 & 0.8300 \\
4 & 0.7100 & 0.9300 & 0.7978 & 0.9103 & 0.8200 \\
5 & 0.8200 & 0.8900 & 0.8497 & 0.8817 & 0.8550 \\
6 & 0.9100 & 0.7900 & 0.8585 & 0.8125 & 0.8500 \\
7 & 0.8600 & 0.8300 & 0.8473 & 0.8350 & 0.8450 \\
8 & 0.8600 & 0.8400 & 0.8515 & 0.8431 & 0.8500 \\
9 & 0.9400 & 0.7200 & 0.8468 & 0.7705 & 0.8300  \\
10 & 0.9300 & 0.8300 & 0.8857 & 0.8455 & 0.8800 \\
\specialrule{0.2pt}{1pt}{1pt}
Ave. & 0.8440 & 0.8430 & 0.8425 & 0.8470 & 0.8435  \\
Std. & 0.0696 & 0.0550 & 0.0259 & 0.0368 & 0.0201  \\
\specialrule{0.3pt}{1pt}{2pt}
\end{tabular}}
\noindent
\justifying{\fontsize{9}{9}\selectfont{Sens. is an abbreviation for Sensitivity, Spec. for Specificity, F1 Sco. for F1 Score, Prec. for Precision, Accu. for Accuracy, Ave. for Average, and Std. for Standard Deviation. These abbreviations are also used in Tables \ref{tab-mixing-patients-results-CNN}, \ref{tab-mixing-patients-results-BiLSTM}, and \ref{tab-cross-patient-results-BiLSTM}.}}
\end{table}

\begin{table}[!htb]
\centering
\fontsize{6}{8}\selectfont
\setlength{\abovecaptionskip}{0pt}
\setlength{\belowcaptionskip}{5pt}
\caption{Cross-validation results using the CNN approach.}
\label{tab-mixing-patients-results-CNN}
\resizebox{\columnwidth}{!}{
\begin{tabular}{l l l l l l}
\specialrule{0.3pt}{1pt}{2pt}
Item & Sens. & Spec. & F1 Sco. & Prec. & Accu. \\
\specialrule{0.2pt}{2pt}{1pt}
1 & 0.8400 & 0.8500 & 0.8442 & 0.8485 & 0.8450 \\
2 & 0.9200 & 0.7700 & 0.8558 & 0.8000 & 0.8450 \\
3 & 0.8000 & 0.8400 & 0.8163 & 0.8333 & 0.8200  \\
4 & 0.9000 & 0.6900 & 0.8145 & 0.7438 & 0.7950  \\
5 & 0.9200 & 0.8000 & 0.8679 & 0.8214 & 0.8600  \\
6 & 0.7900 & 0.8500 & 0.8144 & 0.8404 & 0.8200  \\
7 & 0.6300 & 0.9700 & 0.7590 & 0.9545 & 0.8000 \\
8 & 0.8500 & 0.8700 & 0.8586 & 0.8673 & 0.8600 \\
9 & 0.8700 & 0.7700 & 0.8286 & 0.7909 & 0.8200 \\
10 & 0.9600 & 0.6900 & 0.8458 & 0.7559 & 0.8250 \\
\specialrule{0.2pt}{1pt}{1pt}
Ave. & 0.8480 & 0.8100 & 0.8305 & 0.8256 & 0.8290 \\
Std. & 0.0891 & 0.0809 & 0.0301 & 0.0571 & 0.0217  \\
\specialrule{0.3pt}{1pt}{1pt}
\end{tabular}}
\end{table}
\begin{table}[!htb]
\centering
\fontsize{6}{8}\selectfont
\setlength{\abovecaptionskip}{0pt}
\setlength{\belowcaptionskip}{5pt}
\caption{Cross-validation results using the attention BiLSTM approach.}
\label{tab-mixing-patients-results-BiLSTM}
\resizebox{\columnwidth}{!}{
\begin{tabular}{l l l l l l}
\specialrule{0.3pt}{1pt}{2pt}
Item & Sens. & Spec. & F1 Sco. & Prec. & Accu. \\
\specialrule{0.2pt}{2pt}{1pt}
1 &  0.8800 & 0.9000 & 0.8889 & 0.8980 & 0.8900  \\
2 &  0.8400 & 0.9200 & 0.8750 & 0.9130 & 0.8800  \\
3 &  0.8600 & 0.8400 & 0.8515 & 0.8431 & 0.8500 \\
4 &  0.9400 & 0.7900 & 0.8744 & 0.8174 & 0.8650 \\
5 &  0.9100 & 0.8600 & 0.8878 & 0.8667 & 0.8850\\
6 &  0.8800 & 0.9000 & 0.8889 & 0.8980 & 0.8900\\
7 &  0.8200 & 0.8600 & 0.8367 & 0.8542 & 0.8400\\
8 &  0.8900 & 0.9500 & 0.9175 & 0.9468 & 0.9200\\
9 &  0.8200 & 0.9000 & 0.8542 & 0.8913 & 0.8600\\
10 &  0.8600 & 0.9400 & 0.8958 & 0.9348 & 0.9000\\
\specialrule{0.2pt}{1pt}{1pt}
Ave. & 0.8700  & 0.8860 & 0.8771 & 0.8863 & 0.8780 \\
Std. & 0.0363  & 0.0463 & 0.0228 & 0.0388 & 0.0230 \\
\specialrule{0.3pt}{1pt}{1pt}
\end{tabular}}
\end{table}

For the LSTM approach, the achieved average sensitivity, average specificity and average precision are respectively 84.4\%, 84.3\% and 84.7\%. By using the approach of attention BiLSTM, the obtained average sensitivity of 87\%, specificity of 88.6\% and precision of 88.63\% are better than the LSTM approach. For the F1 score and accuracy, the approach of attention BiLSTM also exceeds the LSTM approach. And the standard deviations of by the attention BiLSTM approach are mostly less than the LSTM approach. It can be seen that the proposed approach of attention BiLSTM not only classifies seizures more accurately than the LSTM approach, but is also more stable.

For the CNN approach, the obtained average sensitivity, average specificity and average precision are 84.8\%, 81.0\% and 82.56\%, respectively. Our model outperforms the CNN approach in sensitivity, specificity and precision. For the average accuracy and the average F1 score, our approach also has higher values than the CNN approach. And the standard deviations in our method are smaller than the CNN approach. These experimental results show that, the proposed approach of attention BiLSTM has better performance in the seizure/non-seizure classification than the CNN approach.

\subsection{Cross-patient seizure/non-seizure classification}
For cross-patient seizure/non-seizure classification, each experiment takes data of one patient as testing data, and other patients' data as training data and validation data according to the ratio 85:15. Because the two cases Chb01 and Chb21 are records from the same patient. The two cases were utilized together either as testing data or training-validation data. In each experiment, all the seizure data segments from each patient were utilized, and non-seizure data segments were randomly selected with the same number of seizure segments. So, the data was balanced in each experiment.

\begin{figure}[ht]
    \centering
    \includegraphics[scale=0.4]{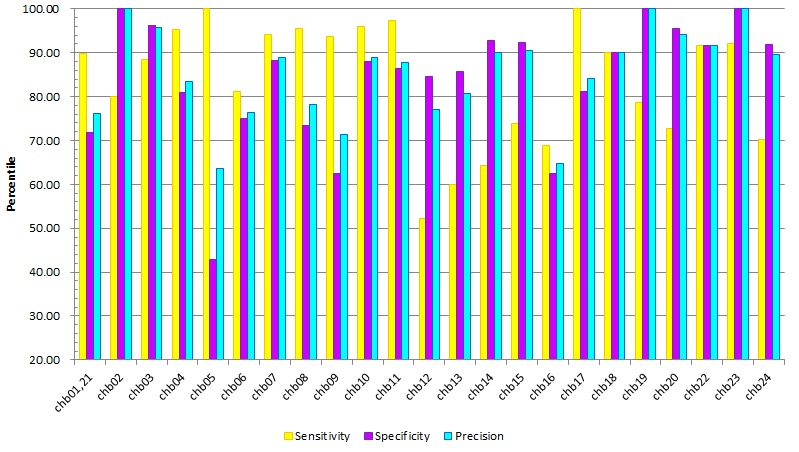}
    \caption{(Color online) Bar chart illustations of cross-patient sensitivity, specificity and precision over 24 cases for the attention BiLSTM.}
    \label{fig-cross-patient-sensitvity-specificity-precision-AttenbiLSTM}
\end{figure}

For each patient, we used her/his EEG data as testing data and data of other patients as training-validation data, and obtained the sensitivity, specificity, F1 score, precision, and accuracy, which are listed in Table \ref{tab-cross-patient-results-BiLSTM}. Fig.  \ref{fig-cross-patient-sensitvity-specificity-precision-AttenbiLSTM} shows the sensitivities, the specificities and the precisions in the form of bar chart. For the 23 patients in CHB-MIT, the average sensitivity, specificity, precision, and accuracy are 83.72\%, 84.06\%, 85.36\%, and 83.89\%, respectively. And the standard deviations of sensitivity, specificity and precision are 0.1349, 0.1379, and 0.1020, respectively.
\begin{table}[!htb]
\small
\centering
\fontsize{6}{8}\selectfont
\setlength{\abovecaptionskip}{0pt}
\setlength{\belowcaptionskip}{5pt}
\caption{Cross-patient experiment results using the attention BiLSTM.}
\label{tab-cross-patient-results-BiLSTM}
\resizebox{\columnwidth}{!}{
\begin{tabular}{l l l l l l}
\specialrule{0.3pt}{1pt}{2pt}
Case & Sens. & Spec. & F1 Sco. & Prec. & Accu.   \\
\specialrule{0.2pt}{2pt}{1pt}
Chb01,21 & 0.8974 & 0.7179 & 0.8235 & 0.7609 & 0.8077  \\
Chb02 & 0.8000 & 1.0000 & 0.8889 & 1.0000 & 0.9000 \\
Chb03 & 0.8846 & 0.9615 & 0.9200 & 0.9583 & 0.9231 \\
Chb04 & 0.9524 & 0.8095 & 0.8889 & 0.8333 & 0.8810 \\
Chb05 & 1.0000 & 0.4286 & 0.7778 & 0.6364 & 0.7143 \\
Chb06 & 0.8125 & 0.7500 & 0.7879 & 0.7647 & 0.7813 \\
Chb07 & 0.9412 & 0.8824 & 0.9143 & 0.8889 & 0.9118 \\
Chb08 & 0.9556 & 0.7333 & 0.8600 & 0.7818 & 0.8444 \\
Chb09 & 0.9375 & 0.6250 & 0.8108 & 0.7143 & 0.7813 \\
Chb10 & 0.9600 & 0.8800 & 0.9231 & 0.8889 & 0.9200 \\
Chb11 & 0.9730 & 0.8649 & 0.9231 & 0.8780 & 0.9189 \\
Chb12 & 0.5211 & 0.8451 & 0.6218 & 0.7708 & 0.6831 \\
Chb13 & 0.6000 & 0.8571 & 0.6885 & 0.8077 & 0.7286 \\
Chb14 & 0.6429 & 0.9286 & 0.7500 & 0.9000 & 0.7857 \\
Chb15 & 0.7379 & 0.9223 & 0.8128 & 0.9048 & 0.8301 \\
Chb16 & 0.6875 & 0.6250 & 0.6667 & 0.6471 & 0.6563 \\
Chb17 & 1.0000 & 0.8125 & 0.9143 & 0.8421 & 0.9063 \\
Chb18 & 0.9000 & 0.9000 & 0.9000 & 0.9000 & 0.9000 \\
Chb19 & 0.7857 & 1.0000 & 0.8800 & 1.0000 & 0.8929 \\
Chb20 & 0.7273 & 0.9545 & 0.8205 & 0.9412 & 0.8409 \\
Chb22 & 0.9167 & 0.9167 & 0.9167 & 0.9167 & 0.9167 \\
Chb23 & 0.9200 & 1.0000 & 0.9583 & 1.0000 & 0.9600 \\
Chb24 & 0.7027 & 0.9189 & 0.7879 & 0.8966 & 0.8108 \\
\specialrule{0.2pt}{1pt}{1pt}
Ave. & 0.8372 & 0.8406 & 0.8363 & 0.8536 & 0.8389 \\
Std. & 0.1349 & 0.1379 & 0.0888 & 0.1020 & 0.0833 \\
\specialrule{0.3pt}{1pt}{1pt}
\end{tabular}}
\end{table}

\begin{figure}[!htb]
    \centering
    \includegraphics[scale=0.47]{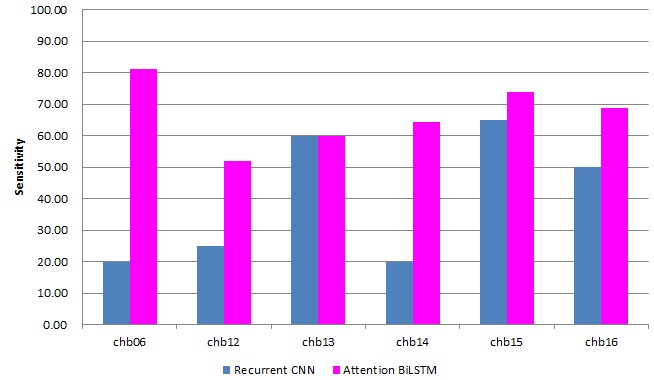}
    \caption{(Color online) Comparison of cross-patient sensitivity over 6 cases between attention BiLSTM and recurrent CNN.}
    \label{fig-comparison-cross-sensitivity-6-cases}
\end{figure}

\begin{figure}[!htb]
    \centering
    \includegraphics[scale=0.47]{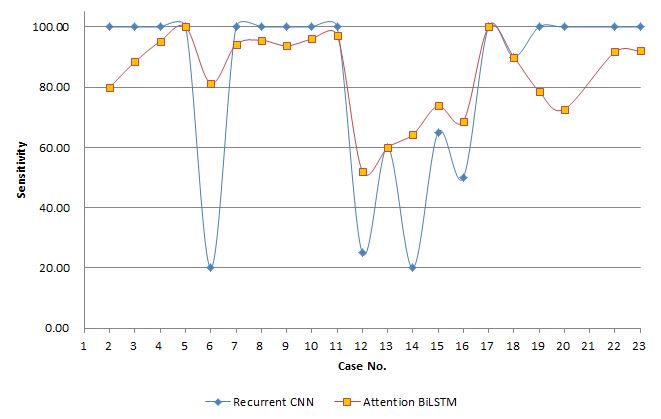}
    \caption{(Color online) Comparison of cross-patient sensitivity over 21 common cases between attention BiLSTM and recurrent CNN.}
    \label{fig-comparison-cross-sensitivity-21-cases}
\end{figure}

In \citep{Thodoroff}, Thodoroff et al. utilize a recurrent convolutional neural network (recurrent CNN) and obtain an average sensitivity 85\% in cross-patient experiments over the CHB-MIT data set. According to Fig. 7(a) and Fig. 7(c) in \citep{Thodoroff}, for six cases Chb06, Chb12, Chb13, Chb14, Chb15 and Chb16, the obtained sensitivity results are not good, only around 20\% for Chb06 and Chb14. For other seventeen cases the sensitivity results are mostly
100\%. The two cases Chb01 and Chb21 are tested separately for recurrent CNN. Our method achieved better sensitivities in the above cases, all exceeding 50\%, although the sensitivity of the remaining cases were less than 100\%.
Fig. \ref{fig-comparison-cross-sensitivity-6-cases} presents the sensitivity comparisons between the method of recurrent CNN and our approach of BiLSTM with attention for the above six cases. And Fig. \ref{fig-comparison-cross-sensitivity-21-cases} shows sensitivities of 21 common-tested cases. The 21 cases do not contain Chb01, Chb21 and Chb24. Over the common-tested cases, our standard deviations for sensitivity and specificity are 0.1374 and 0.1407, respectively. The results indicated that our sensitivity results are more concentrative, and in this sense, the proposed approach of attention BiLSTM is more stable.

\section{Model analysis}
\label{section-model-analysis}

\subsection{Interpretations of attention mechanism}
\begin{figure}[ht]
\centering
\includegraphics[scale=0.35]{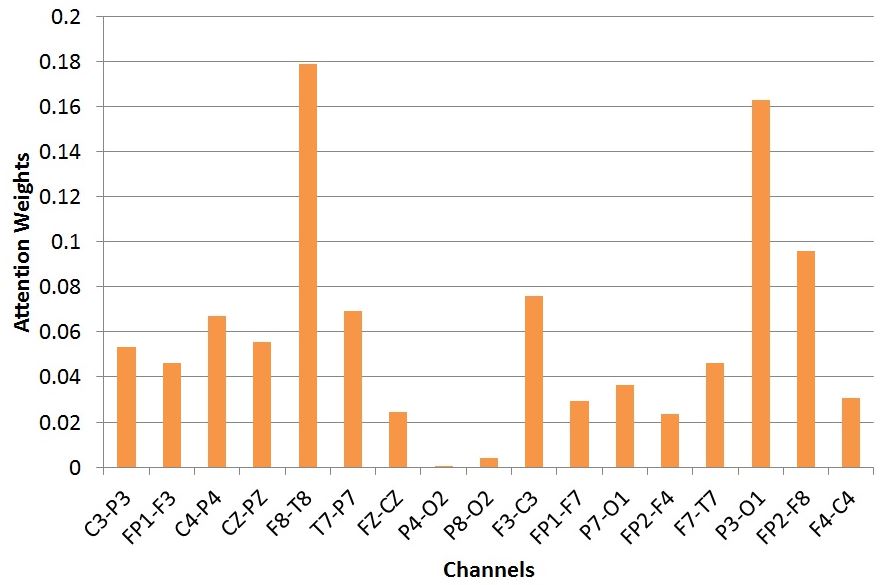}
\caption{Attention weights on channels for a seizure segment in Chb11.}
\label{attention-weights-for-seizure-segment-in-chb11}
\end{figure}

\begin{figure}[ht]
\centering
\includegraphics[scale=0.35]{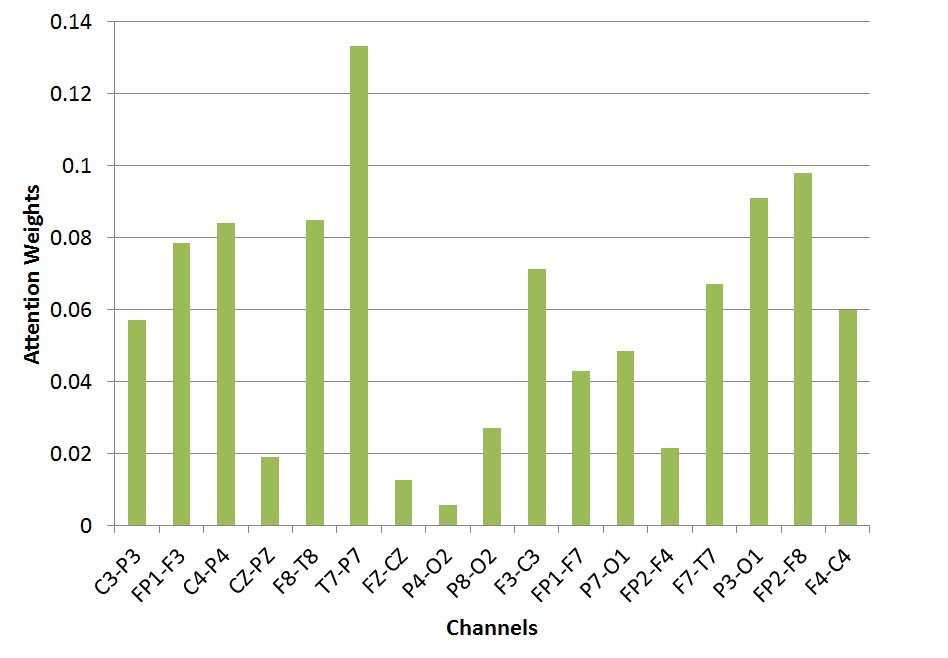}
\caption{Attention weights on channels for a seizure segment in Chb03.}
\label{attention-weights-for-seizure-segment-in-chb03}
\end{figure}

Our attention mechanism is designed for distinguishing signals from different brain regions and produces different weights for the signals. In the attention layer, a kernel matrix and a bias matrix are needed, and they are trained together with other modules in our model. Based on the two matrices, the weights of channels, which correspond to different brain regions, are calculated according to the input data.
In fact, different epilepsy patients have different seizure patterns and EEG signal is dynamic. For one patient, experienced seizures may have different types and may come from different brain regions. Therefore, it is reasonable to calculate adaptively channel weights in our attention mechanism.
Fig. \ref{attention-weights-for-seizure-segment-in-chb11} and Fig. \ref{attention-weights-for-seizure-segment-in-chb03} show attention weight distributions on 17 channels in two data segments from two patients (i.e., Chb11 and Chb03), which are computed by the attention mechanism in the same trained model. These two figures show that our attention mechanism can adaptively calculate the channel weights of signal data from different patients.

\begin{figure}[!htb]
\centering
\subfigure[Signals in a non-seizure segment from Chb11.]{
\label{fig-atten-weights-chb11-non-seizure-a}
\includegraphics[scale=0.37]{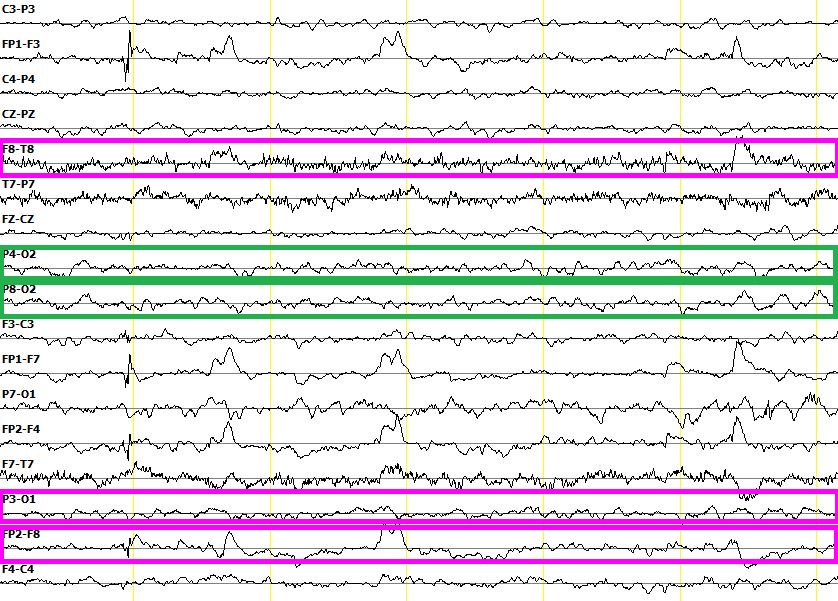}
}
\vspace{0.5cm}
\subfigure[Signals in a seizure segment from Chb11.]{
\label{fig-atten-weights-chb11-seizure-b}
\includegraphics[scale=0.37]{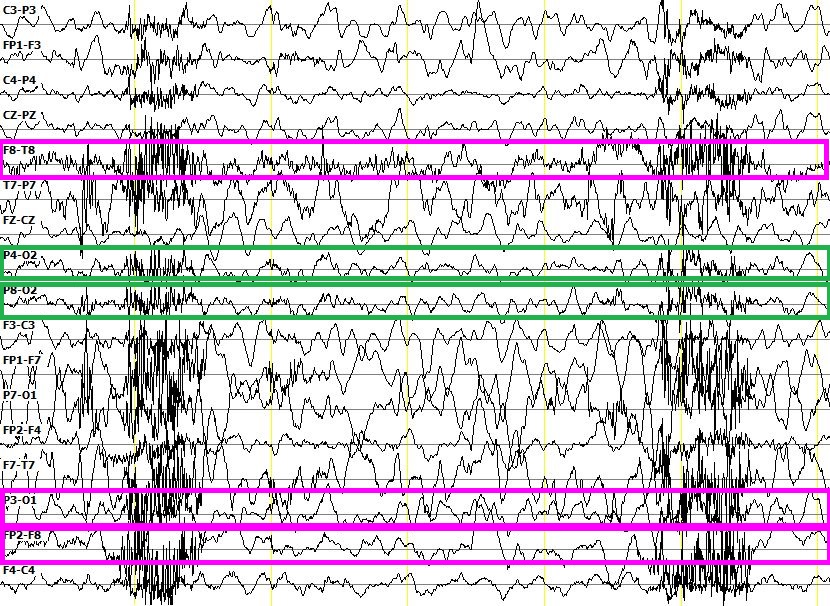}
}
\vspace{-0.7cm}
\caption{(Color online) Visualization of signals on channels in a non-seizure segment and a seizure segment from Chb11. Purple panels represent channels with large signal changes, and green panels for channels with small signal changes.}
\label{seizure-nonseizure-slices-from-chb11}
\end{figure}

In some areas of the brain, EEG signals during seizures show many differences with signals at non-seizures. The differences, such as  frequency and magnitude, could be used to indentify seizure and non-seizure. The attention mechanism captures signal characteristics and assigns large weight values to the channels, which could distinguish seizure and non-seizure segments. Generally, the greater the differences between the seizure signal and the non-seizure signal through the channel, the greater the weight assigned to the corresponding channel. An example of attention weights of 17 channels for a seizure segment is shown in Fig. \ref{attention-weights-for-seizure-segment-in-chb11}; the channels of F8-T8, P3-O1 and FP2-F8 have the large weights compared to other channels. In Fig. \ref{fig-atten-weights-chb11-non-seizure-a} and Fig. \ref{fig-atten-weights-chb11-seizure-b}, the actual signals over the above three channels change (i.e., six purple panels) much either in the frequency or in the magnitude. For the actual signals over channels P4-O2 and P8-O2 (i.e., four green panels), the changes in Fig. \ref{fig-atten-weights-chb11-non-seizure-a} and Fig. \ref{fig-atten-weights-chb11-seizure-b} are relatively small. As shown in Fig. \ref{attention-weights-for-seizure-segment-in-chb11}, the assigned weights over the two channels are small.

\begin{figure}[!htb]
\addtocounter{subfigure}{2}
\centering
\subfigure[Signals in a non-seizure segment from Chb03.]{
\label{fig-atten-weights-chb03-02-non-a}
\includegraphics[scale=0.37]{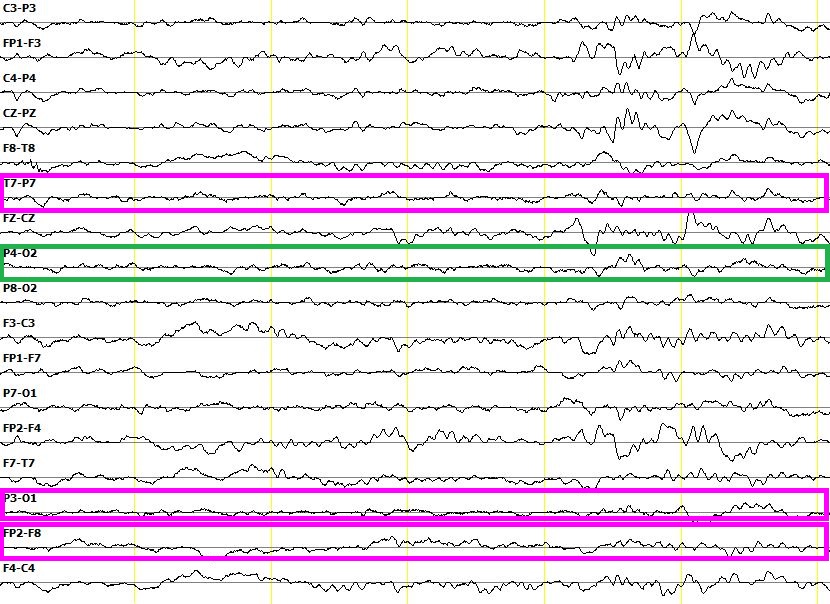}
}
\vspace{0.5cm}
\subfigure[Signals in a seizure segment from Chb03.]{
\label{fig-atten-weights-chb03-02-b}
\includegraphics[scale=0.37]{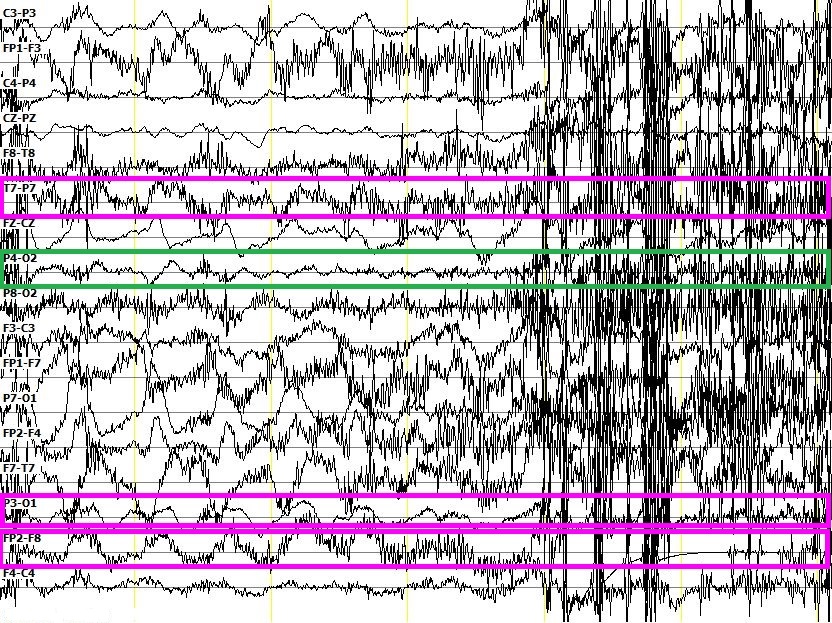}
}
\vspace{-0.7cm}
\caption{(Color online) Visualization of signals on channels in a non-seizure segment and a seizure segment from Chb03. The purple panels and green panels have the same meanings as in Fig. \ref{seizure-nonseizure-slices-from-chb11}. }
\label{seizure-nonseizure-slices-from-chb03}
\end{figure}

The actual signals of the channel P4-O2 (i.e., two green panels) in Fig. \ref{fig-atten-weights-chb03-02-non-a} and Fig. \ref{fig-atten-weights-chb03-02-b} manifest small differences in magnitudes. The attention mechanism produces small weight for the channel P4-O2 so that the corresponding signal data is not treated as critical evidences to classify seizure/non-seizure. The signals over channels T7-P7, FP2-F8 and P3-O1 (i.e., six purple panels) change a lot from the non-seizure Fig. \ref{fig-atten-weights-chb03-02-non-a} to the seizure Fig. \ref{fig-atten-weights-chb03-02-b}. Such changes could differentiate seizure/non-seizure segments. So, the three channels are assigned large attention weights, as shown in Fig. \ref{attention-weights-for-seizure-segment-in-chb03}.

\subsection{Validations of BiLSTM and attention mechanism}
The approach of attention BiLSTM is developed in the inspiration of the LSTM approach in \citep{Hussein}. In the development, the performances of  bidirectional LSTM and attention mechanism are separately explored. By using parameters with the best performances in the tuning procedures, ten rounds of cross-validation experiments are performed separately for testing the two modules.
When testing the module of bidirectional LSTM, the parameters are set as follows: The learning rate is 0.001, the number of hidden states in the bidirectional LSTM is 100, that in time-distributed layer is 50, the optimizer is RMSprop, batch size is 30, and the number of epochs is 30. For the testing of attention mechanism, the parameters are: The learning rate is 0.001, the number of hidden states in the module LSTM is 100, that in time-distributed layer is 50, the optimizer is RMSprop, batch size is 30, and the number of epochs is 25. The obtained cross-validation results are shown in Table \ref{tab-results-of-module-in-atten-BiLSTM}. The results indicate that, the bidirectional LSTM only improves the sensitivity and the attention mechanism only enhances the specificity compared to the LSTM approach results in Table \ref{tab-mixing-patients-results-LSTM}. After combining the two modules in the approach of attention BiLSTM, both the sensitivity and the specificity are improved with 2.6\% and 4.3\%, respectively. Thus, both bidirectional LSTM and attention mechanism play important roles in the approach of attention BiLSTM for seizure/non-seizure classification.
\begin{table*}[!htb]
\centering
\fontsize{3}{4}\selectfont
\setlength{\abovecaptionskip}{0pt}
\setlength{\belowcaptionskip}{10pt}
\caption{Cross-validation results for modules in the attention BiLSTM approach.}
\label{tab-results-of-module-in-atten-BiLSTM}
\resizebox{\textwidth}{!}{
\begin{tabular}{l l l l l l}
\specialrule{0.2pt}{0.5pt}{1pt}
Module & Sensitivity & Specificity & F1 Score & Precision & Accuracy \\
\specialrule{0.1pt}{1pt}{0.5pt}
Bidirectional LSTM & 0.8630$\pm$0.06 & 0.8280$\pm$0.05 & 0.8477$\pm$0.01 & 0.8373$\pm$0.03 & 0.8455$\pm$0.01 \\
Attention LSTM & 0.8340$\pm$0.05 & 0.8870$\pm$0.04 & 0.8564$\pm$0.02 & 0.8828$\pm$0.03 & 0.8605$\pm$0.02   \\
Attention BiLSTM & 0.8700$\pm$0.04 & 0.8860$\pm$0.05 & 0.8771$\pm$0.02 & 0.8863$\pm$0.04 & 0.8780$\pm$0.02 \\
\specialrule{0.2pt}{0.5pt}{0.5pt}
\end{tabular}}
\end{table*}

\section{Discussion}
\label{section-discussion}
In this paper, we design a novel approach of BiLSTM with attention for seizure/non-seizure classification in off-line EEG data. Cross-patient and cross-validation experiments across patients are separately applied to evaluations on the pediatric data set of CHB-MIT. When doing segmentation, a time length of 23 seconds is selected by referring to the segment length in Bonn EEG data set \citep{Andrzejak}, and each data record in each case is split from the beginning to the end without overlapping. As a result, 665 seizure segments are obtained, and the lengths of seizure data vary from 1s to 23s in seizure segments. The length diversity of seizure data is aligned with a real-world situation. In each experiment, the 665 seizure segments were taken as a part of experimental data, and 665 non-seizure segments were randomly selected from the extracted non-seizure segments. Its randomness and sparsity reduce temporal correlations among non-seizure data segments, and avoid resulting in overly optimistic specificity results \citep{Shoeb}. The above segmenting-data-record way and the selecting strategy of non-seizure segments make the evaluation of our approach be more reliable.

In the cross-validation experiments, the sensitivity, specificity and precision of our approach were better than the LSTM approach in \citep{Hussein} and the CNN approach in \citep{Acharya}. The improvements in the sensitivity, specificity, and precision over those two state-of-the-art approaches were 2.6\%, 4.3\%, 3.93\% and 2.2\%, 7.6\%, 6.07\%, respectively, and the standard deviations were less than the two approaches in comparison. As Table \ref{tab-results-of-module-in-atten-BiLSTM} shows, the better performances of our approach are attributed to the attention mechanism and the feature extraction in both forward and backward directions.

Among cross-patient experiment results in Table \ref{tab-cross-patient-results-BiLSTM}, there exist gaps. Over the six patients, including Chb05, Chb09, Chb12, Chb13, Chb14, and Chb16, either sensitivity or specificity were less than 70\%. For the seven patients, i.e., Chb03, Chb07, Chb10, Chb11, Chb18, Chb22 and Chb23, all testing results were over 85\%. The possible reason is that, for a child, the brain, meninges, skull, and head size change overtime \citep{Emory_University}. Compared to the method of recurrent CNN proposed in \citep{Thodoroff}, the performances of our method BiLSTM with attention were more stable. In \citep{Thodoroff}, the convolution neural network module in recurrent CNN is pre-trained before training the whole model. Our attention BiLSTM approach does not need pre-training, and it directly processes raw data and extracts features. The REVEAL algorithm proposed in \citep{Wilson} achieved an average sensitivity of 61\%. \citep{Furbass} used the automatic seizure detection system EpiScan on the CHB-MIT data set and obtained an average sensitivity of 67\%. The average sensitivity of our approach is much better than REVEAL and EpiScan.

The application scenario of our approach is to automatically selecting all the seizure segments from the off-line EEG data records for neurologists' reviewing and analysis. Because of the off-line EEG data segments, extracting features in the forward direction and the backward direction and performing analyses are feasible in practices. In the application, selecting as many seizure segments as possible and as accurately as possible is our target. For this target, the performance of a seizure-segments-selection method can be measured by sensitivity, specificity and precision, not by the number of false alarms per hour. So, the metric of false alarm rate is not calculated and not compared for the proposed approach.

Instead of directly training weights on channels, we utilize an attention mechanism to generate weights. In the directly training way, the obtained weights on channels are the same for all the patients. In fact, the seizure patterns of different patients are different, and different types of seizures have different patterns, and it is possible that one patient may have different types of seizures. Therefore, for data segments from different patients, the weights on channels, which describe the strength that signals signify seizures, need be different. In our attention mechanism, a kernel matrix and a bias matrix are obtained by training, and then the two trained matrices are performed transformations by combining with data segments. The outputs of  transformations are attention weights for the data segments. The attention mechanism produce different weights for data segments from different patients, and further efficiently help extract seizure features.

When designing attention mechanism, we tried different ways: one way is adding different attention weights over time steps, and another way is adding different attention weights over time steps and over channels. Our experimental results using the two ways were not good. One possible reason is that the role of each brain region in the whole brain state is generally stable in a short duration such as 23s. Finally, we choose to apply attention mechanism to channels and share the attention weights among time steps. Actually, different channels have different contributions to a seizure, and the contributions turn out to be correlated to the locations of brain regions, rather than the time. In addition, we applied our method to single channel data. The results with single channel data were not good. They were in agreement with the observation in \citep{Shoeb}; that is, for some channels, the data morphology in seizure state is similar to that in non-seizure state.

\section{Conclusions}
\label{section-conclusion}
This paper focuses on the problem of automatic seizure/non-seizure classification. Inspired by the architecture in \citep{Hussein}, we analyze both spacial and temporal characteristics of seizures, and propose a novel deep learning-based approach by using the model of BiLSTM integrated with attention. The integration of an attention mechanism is to capture spatial features better, and the employment of the BiLSTM model is to extract more temporal features. The proposed approach is evaluated on the noisy EEG data set of CHB-MIT. The evaluation is across multiple patients and uses data from multiple brain regions. In the cross-validation experiments, we obtain sensitivity of 87\%, specificity of 88.6\% and precision of 88.63\%, which are better than the LSTM approach in \citep{Hussein} and the CNN approach in \citep{Acharya}. In the cross-patient experiments, the testing results are 83.72\%-sensitivity, 84.06\%-specificity and 85.35\%-precision on average. Comparing to the model reccurrent CNN in \citep{Thodoroff}, our model BiLSTM with attention is more stable.

In the approach of BiLSTM with attention, the pooling layer adopts a globally-averaging way to extract holistic features of data segments. The problem whether such a way is the best or not for the seizure/non-seizure classification will be explored in the future. And also we want to investigate whether the length of data segments has effects on the sensitivity, the specificity and the precision.




\begin{thebibliography}{1}


\bibitem{Megiddo}
I. Megiddo, A. Colson, D. Chisholm, T. Dua, A. Nandi, R. Laxminarayan. Health and economic benefits of public financing of epilepsy treatment in India: An agent-based simulation model. Epilepsia, vol. 57, no. 3, pp. 464-474, 2016.  https://doi.org/10.1111/epi.13294.

\bibitem{Gotman1979}
J. Gotman, J. R. Ives, P. Gloor. Automatic recognition of inter-ictal epileptic activity in prolonged EEG recordings. Electroencephalography and Clinical Neurophysiology, vol. 46, no. 5, pp. 510-520, 1979. https://doi.org/10.1016/0013-4694(79)90004-X.

\bibitem{Gotman1982}
J. Gotman. Automatic recognition of epileptic seizures in the EEG. Electroencephalography and Clinical Neurophysiology, vol. 54, no. 5, pp. 530-540, 1982. https://doi.org/10.1016/0013-4694(82)90038-4.

\bibitem{Thodoroff}
P. Thodoroff, J. Pineau, A. Lim. Learning robust features using deep learning for automatic seizure detection. Proceedings of the 1st Machine Learning for Healthcare Conference; Los Angeles, CA, USA; 2016. Journal of Machine Learning Research, vol. 56, pp. 178-190, 2016.

\bibitem{Furbass}
F. F\"{u}rbass, P. Ossenblok, M. Hartmann, H. Perko, A. M. Skupch, G. Lindinger, L. Elezi, E. Pataraia, A. J. Colon, C. Baumgartner, T. Kluge. Prospective multi-center study of an automatic online seizure detection system for epilepsy monitoring units. Clinical Neurophysiology, vol. 126, no. 6, pp. 1124-1131, 2015. https://doi.org/10.1016/j.clinph.2014.09.023.

\bibitem{Zandi}
A. S. Zandi, M. Javidan, G. A. Dumont, R. Tafreshi. Automated real-time epileptic seizure detection in scalp EEG recordings using
an algorithm based on wavelet packet transform. IEEE Transactions on Biomedical Engineering, vol. 57, no. 7, pp.1639-1651, 2010. https://doi.org/10.1109/TBME.2010.2046417.

\bibitem{Shoeb}
A. Shoeb, J. Guttag. Application of machine learning to epileptic seizure detection. Proceedings of the 27th International Conference on Machine Learning; pp. 975-982; Haifa, Israel; 2010.

\bibitem{Amin}
S. Amin, A. M. Kamboh. A robust approach towards epileptic seizure detection. Proceedings of IEEEE 26th International Workshop on Machine Learning for Signal Processing; pp. 1-6; Salerno, Italy; 2016.

\bibitem{Fan}
M. Fan, C. Chou. Detecting abnormal pattern of epileptic seizures via temporal synchronization of EEG signals. IEEE Transactions on Biomedical Engineering, vol. 66, no. 3, pp. 601-608, 2019. https://doi.org/10.1109/TBME.2018.2850959.

\bibitem{Hunyadi}
B. Hunyadi, M. Signoretto, W. V. Paesschen, J. A. K. Suykens, S. V. Huffel, M. D. Vos. Incorporating structural information from the multichannel EEG improves patient-specific seizure detection. Clinical Neurophysiology, vol. 123, no. 12, pp. 2352-2361, 2012. https://doi.org/10.1016/j.clinph.2012.05.018.

\bibitem{Esbroeck}
A. V. Esbroeck, L. Smith, Z. Syed, S. Singh, Z. Karam. Multitask seizure detection: Addressing intra-patient variation in seizure
morphologies. Machine Learning, vol. 102, no. 3, pp. 309-321, 2016. https://doi.org/10.1007/s10994-015-5519-7.

\bibitem{Truong2017}
N. D. Truong, L. Kuhlmann, M. R. Bonyadi, J. Yang. Supervised learning in automatic channel selection for epileptic seizure detection. Expert Systems with Applications, vol. 86, pp. 199-207, 2017. https://doi.org/10.1016/j.eswa.2017.05.055.

\bibitem{Vidyaratne}
L. Vidyaratne, A. Glandon, M. Alam, K. M. Iftekharuddin. Deep recurrent neural network for seizure detection. Proceedings of 2016 International Joint Conference on Neural Networks; pp. 1202-1207; Vancouver, BC, Canada; 2016.


\bibitem{Qu}
H. Qu, J. Gotman. Improvement in seizure detection performance by automatic adaptation to the EEG of each patient. Electroencephalography and Clinical Neurophysiology, vol. 86, no. 2, pp. 79-87, 1993. https://doi.org/10.1016/0013-4694(93)90079-B.

\bibitem{Zhou}
W. Zhou, Y. Liu, Q. Yuan, X. Li. Epileptic seizure detection using lacunarity and Bayesian linear discriminant analysis in intracranial EEG. IEEE Transactions on Biomedical Engineering, vol. 60, no. 12, pp. 3375-3381, 2013. https://doi.org/ 10.1109/TBME.2013.2254486.

\bibitem{Golmohammadi}
M. Golmohammadi, S. Ziyabari, V. Shah, E. V. Weltin, C. Campbell, L. Obeid, J. Picone. Gated recurrent networks for seizure detection. Proceedings of the IEEE Signal Processing in Medicine and Biology Symposium; pp. 1-5; Philadelphia, Pennsylvania, USA; 2017.

\bibitem{Acharya}
U. R. Acharya, S. L. Oh, Y. Hagiwara, J. H. Tan, H. Adeli. Deep convolutional neural network for the automated detection and
diagnosis of seizure using EEG signals. Computers in Biology and Medicine, vol. 100, no. 1, pp. 270-278, 2018. https://doi.org/10.1016/j.compbiomed.2017.09.017.

\bibitem{Hussein}
R. Hussein, H. Palangi, R. Ward, Z. J. Wang. Epileptic seizure detection: A deep learning approach. arXiv prepint, arXiv:1803.09848v1, 2018.

\bibitem{Fergus}
P. Fergus, A. Hussain, D. Hignett, D. A. Jumeily, K. A. Aziz, H. Hamdan. A machine learning system for automated whole-brain seizure
detection. Applied Computing and Informatics, vol. 12, no. 1, pp. 70-89, 2016. https://doi.org/10.1016/j.aci.2015.01.001.

\bibitem{Nicalaou}
N. Nicalaou, J. Georgiou. Detection of epileptic electroencephalogram based on permutation entropy and support vector machines.
 Expert Systems with Applications, vol. 39, no. 1, pp. 202-209, 2012. https://doi.org/10.1016/j.eswa.2011.07.008.

\bibitem{Nasehi}
S. Nasehi, H. Pourghassem. Patient-specific epileptic seizure onset detection algorithm based on spectral features and IPSONN classifier. Proceedings of 2013 International Conference on Communication Systems and Network Technologies; pp. 186-190; Gwalior, India; 2013.

\bibitem{Kharbouch}
A. Kharbouch, A. Shoeb, J. Guttag, S. S. Cash. An algorithm for seizure onset detection using intracranial EEG. Epilepsy \& Behavior, vol. 22, no. 1, pp. S29-S35, 2011. https://doi.org/10.1016/j.yebeh.2011.08.031.

\bibitem{Zheng}
Y. Zheng, J. Zhu, Y. Qi, X. Zheng, J. Zhang. An automatic patient-specific seizure onset detection method using intracranial electrocephalography. Neuromodulation, vol. 18, no. 2, pp. 79-84, 2015. https://doi.org/10.1111/ner.12214.

\bibitem{Acharya2012}
U. R. Acharya, F. Molinari, S. V. Sree, S. Chattopadhyay, K.-H. Ng, J. S.Suri. Automated diagnosis of epileptic EEG using entropies. Biomedical Signal Processing and Control, vol. 7, no. 4, pp. 401-408, 2012. https://doi.org/10.1016/j.bspc.2011.07.007.

\bibitem{Ansari}
A. H. Ansari, P. J. Cherian, A. Caicedo, G. Naulaers, M. Vos, S. V. Huffel. Neonatal seizure detection using deep convolutional neural networks. International Journal of Neural Systems, vol. 28, no. 0, 1850011, 2018. https://doi.org/10.1142/S0129065718500119.

\bibitem{Kiranyaz}
S. Kiranyaz, T. Ince, M. Zabihi, D. Ince. Automated patient-specific classification of long-term electroencephalography. Journal of Biomedical Informatics, vol. 49, pp. 16-31, 2014. https://doi.org/10.1016/j.jbi.2014.02.005.



\bibitem{Hassanpour}
H. Hassanpour, M. Shahiri. Adaptive segmentation using wavelet transform. Proceedings of 2007 International Conference on Electrical Engineering; pp.1-5; Lahore, Pakistan; 2007.

\bibitem{Schuster}
M. Schuster, K. K. Paliwal. Bidirectional recurrent neural networks. IEEE Transactions on Signal Processing, vol. 45, no. 11, pp. 2673-2681, 1997. https://doi.org/10.1109/78.650093.

\bibitem{Graves}
A. Graves, J. Schmidhuber. Framewise phoneme classification with bidirectional LSTM and other neural network architectures. Neural Networks, vol. 18, no. 5-6, pp. 602-610, 2005. https://doi.org/10.1016/j.neunet.2005.06.042.

\bibitem{Gers}
F. A. Gers, J. Schmidhuber, F. Cummins. Learning to forget: Continual prediction with LSTM. Neural Computation, vol. 12, no. 10, pp. 2451-2471, 2000. https://doi.org/10.1162/089976600300015015.

\bibitem{Greff}
K. Greff, R. K. Srivastava, J. Koutn\'{i}k, B. R. Steunebrink, J. Schmidhuber. LSTM: A search space odyssey. IEEE Transactions on Neural Networks and Learning Systems, vol. 28, no. 10, pp. 2222-2232, 2017. https://doi.org/10.1109/TNNLS.2016.2582924.

\bibitem{Andrzejak}
R. G. Andrzejak, K. Lehnertz, F. Mormann, C. Rieke, P. David, C. E. Elger. Indications of nonlinear deterministic and finite-dimensional structures in time series of brain electrical activity: Dependence on recording region and brain state. Physical Review E, vol. 64, 061907, 2001. https://doi.org/10.1103/PhysRevE.64.061907.

\bibitem{Emory_University}
P. J. Holt. Introduction to pediatric EEG. Atlanta: Emory University School of Medicine; c2017. Available from: https://www.pediatrics.emory.edu/divisions/neurology/education/pedeeg.html

\bibitem{Wilson}
S. B. Wilson, M. L. Scheuer, R. G. Emerson, A. J. Gabor. Seizure detection: Evaluation of the Reveal algorithm. Clinical Neurophysiology, vol. 115, no. 10, pp. 2280-2291, 2004. https://doi.org/10.1016/j.clinph.2004.05.018.


\end{thebibliography}



\end{document}